# Prompt Chaining in Practice: A Case Study in Automated Scholarly Report Generation

Andrei Lazarev
Department of Control and Applied Mathematics
Moscow Institute of Physics and Technology
Moscow, Russia
ORCID:0009-0008-1519-5820

*Abstract*—**The exponential growth of scholarly publications requires automated tools for effective information synthesis. However, simple, single-shot prompting methods often lack the reliability and quality required for complex synthesis tasks. This paper introduces and empirically evaluates a multi-stage prompt chaining methodology as a more reliable architectural pattern for such tasks. This approach is implemented in our system, AI SciBrief, which automatically generates scholarly digests. We conducted a comparative experiment, measuring the performance of our prompt chaining method against a carefully optimized single-shot baseline. Both systems were evaluated against a human-authored "gold standard" report for the "Education" domain. The results demonstrate a significant difference in reliability: our prompt chaining method achieved a 100% success rate, whereas the optimized baseline failed in 50% of its runs. In terms of quality, the proposed method also demonstrated a clear advantage, achieving a superior ROUGE-L F1-score (0.507 vs. 0.486), driven primarily by higher precision. We conclude that prompt chaining is a more dependable and effective engineering approach for complex, multi-step generative tasks, significantly mitigating the risks of failure and inconsistency inherent in monolithic prompts.**

## I. INTRODUCTION

The modern scientific landscape is characterized by an exponential growth of scholarly output, a phenomenon well-documented in bibliometric studies [1]. With estimates suggesting that thousands of new papers are published daily, researchers face a significant challenge of information overload [2]. This rapidly increasing volume of publications poses a substantial obstacle to staying current, identifying emerging trends, and discovering opportunities for innovation. While this issue affects all researchers, it poses a particular challenge in fast-moving, data-intensive fields such as finance, medicine and artificial intelligence.

Large Language Models (LLMs) have emerged as a promising technology to address this challenge. Their ability to process and synthesize vast amounts of unstructured text offers a powerful alternative to traditional, manual methods of literature review. However, early approaches often rely on naive, single-shot prompting, which, when applied to complex synthesis tasks, can lead to superficial outputs, loss of critical context, and factual inaccuracies or "hallucinations" [3]. To unlock the full potential of LLMs for high-quality analysis, more sophisticated methods are required.

One of the most effective advanced techniques is "prompt chaining", a method that decomposes a single complex query into a sequence of simpler, interconnected prompts. This approach mirrors a human cognitive process, where a large problem is broken down into manageable steps. By guiding the LLM through a logical chain of reasoning - from thematic clustering to trend analysis and final synthesis - it is possible to achieve a significantly higher degree of accuracy, coherence, and depth, building upon foundational concepts like Chain-of-Thought prompting [4].

This paper presents the first detailed technical implementation and quantitative evaluation of a prompt chaining approach, implemented in our system, AI SciBrief (https://sci-brief.com) [5]. The foundational concept of AI SciBrief as a tool for financial analysis was introduced in our prior work [6], [7]. This paper moves significantly beyond that initial concept to focus on the core methodological improvements. The primary contributions of this work are threefold:

*1)* We present a detailed, three-stage **architecture of a prompt chaining pipeline** designed specifically for the task of automated scholarly digest generation.

*2)* We provide a comprehensive empirical performance evaluation of the system, presenting quantitative metrics for its **quality (ROUGE)** and **reliability (Success Rate)**.

## II. PROBLEM SETUP

To move from the mere collection of academic literature to its effective synthesis, a solution must address two distinct challenges. First, it must be capable of performing high-quality **automated scientific summarization**, moving beyond simple keyword extraction to genuine synthesis. Second, to achieve this quality, it must employ **sophisticated prompt engineering techniques** to reliably control the powerful but complex behavior of Large Language Models. This section reviews the **recent advancements** in both areas, establishing the context and rationale for our chosen approach.

### *A. Automated Scientific Summarization*

Early approaches to automated text summarization primarily focused on extractive methods, which identify and concatenate the most salient sentences from a source text. While effective for single documents, these methods struggle with the more complex task of synthesizing information from multiple sources. Foundational work in the field focused on creating the necessary infrastructure, such as large-scale, machine-readable corpora of scientific papers like S2ORC, which paved the way for more advanced analytical techniques [8].

The advent of Large Language Models has fundamentally shifted the landscape, making high-quality abstractive summarization a practical reality: the model generates new text to summarize the source material. Comprehensive surveys now recognize the potential of LLMs to assist researchers with literature reviews, positioning them as powerful tools for accelerating the research process while also cautioning against challenges like factual accuracy and inherent bias [9]. However, a key challenge remains: generating a coherent, structured, and insightful summary from a large, heterogeneous collection of documents is a task that goes beyond simple text compression. It requires a system to identify overarching themes, synthesize disparate findings, and present them in a logical narrative, a process that is not reliably achieved with naive, single-shot prompting.

### *B. Prompt Engineering Techniques*

The effectiveness of an LLM is highly dependent on the quality of the instructions, or prompts, it receives. Consequently, the field of prompt engineering has rapidly evolved from simple instructions to a variety of complex, structured techniques designed to elicit more sophisticated reasoning and behavior from models [10].

A significant breakthrough in this area was **Chain-of-Thought (CoT) prompting**, which demonstrated that instructing a model to "think step-by-step" before providing a final answer dramatically improves its performance on complex reasoning tasks [4]. This discovery proved that the quality of an LLM's output is directly related to the cognitive process it is guided to simulate. This has inspired the development of even more advanced reasoning structures, such as Tree-of-Thought (ToT), which explores multiple reasoning paths in parallel.

This work employs a related but distinct technique known as **Prompt Chaining**. Unlike CoT, which typically focuses on structuring the reasoning process within a single prompt-response cycle, prompt chaining coordinates a workflow across **multiple, sequential prompt-response interactions**. In this approach, the output of one prompt becomes the structured input for the next, creating a pipeline that systematically transforms raw data into a final, synthesized product.

This approach is particularly well-suited for the task of automated digest generation, as the task itself is inherently multi-stage: it requires thematic grouping, intra-group summarization, and final narrative assembly. Therefore, this paper presents **a case study on the practical application and performance of a prompt chaining methodology** designed to solve this specific problem. We detail **the architecture of the prompt chain** itself and provide empirical results to validate its effectiveness.

## III. Prompt Chaining Pipeline

Our implementation, AI SciBrief, operationalizes the prompt chaining methodology through a carefully designed three-stage pipeline. The foundational concept of using an LLM to generate financial digests, including a high-level overview of a three-stage process, was first introduced in our prior work [7]. This paper significantly expands upon that initial framework by providing a detailed, domain-agnostic description of the prompt engineering at each stage. The core principle is to deconstruct the complex, abstract task of "summarizing a field" into a sequence of concrete, manageable sub-tasks. This approach extends the foundational principle of Chain-of-Thought (CoT) prompting, which enhances reasoning within a single turn [4], to a multi-turn workflow, where the structured output of one stage becomes the input for the next. This creates a reliable and controllable process that significantly improves the quality and coherence of the final output. The overall architecture of this pipeline is illustrated in **Figure 1**.

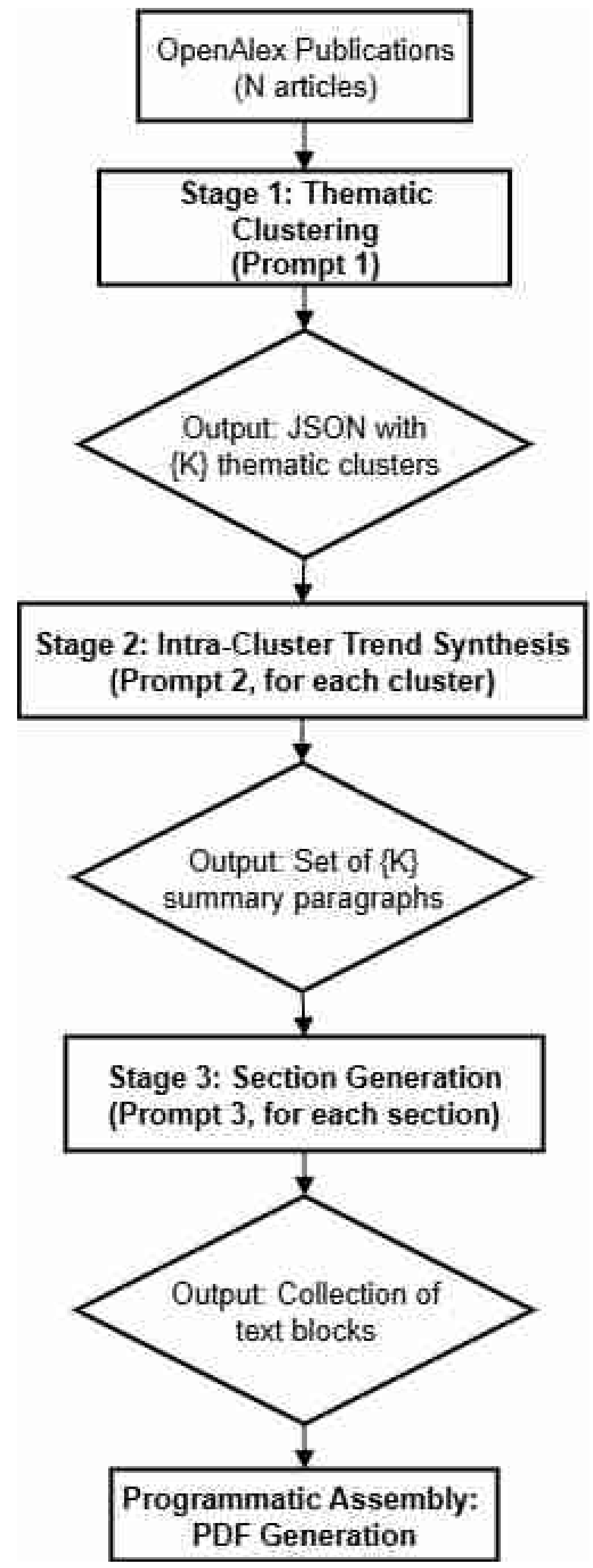

Fig. 1. The three-stage prompt chaining pipeline of the AI SciBrief system.

The following subsections detail the objective, implementation, and data transformation at each stage of the pipeline.

### *A. Stage 1: Thematic Clustering*

*Objective:* The initial challenge is to extract relevant thematic signals of thousands of individual publications. The objective of this stage is to move from a flat list of articles to a structured, thematic grouping by identifying the most dominant research topics discussed within the dataset for a given month.

*Methodology:* At this stage, we leverage the LLM's advanced capability for pattern recognition and topic modeling. We construct a specific prompt that instructs the model to act as a research analyst. The model is fed the titles and abstracts of all collected publications for a specific domain. The prompt explicitly asks the model to identify a predefined number of the most significant, recurring themes and to return the result in a machine-readable JSON format. The number of themes, {K}, is a configurable parameter; for the purposes of this study and our monthly digests, it was empirically set to 5 for this study to ensure a balance between breadth and depth.

**Prompt 1 (Pseudo-code):**

ROLE: You are an expert research analyst specializing in [Domain].

CONTEXT: You are provided with a list of {N} academic paper titles and abstracts published in the last month.

TASK:

1. Read and analyze all provided texts.

2. Identify the {K=5} most dominant and distinct research themes that emerge from this collection.

3. For each theme, provide a short, descriptive title (e.g., "AI in ESG Scoring").

4. Group the original paper IDs under the theme they best represent.

OUTPUT FORMAT: Return your response ONLY as a valid JSON object with the following structure:

```
{
"themes": [
 {
  "theme_title": "...",
  "paper_ids": ["id1", "id2", ...]
  },
 ...
 ]
}
```

**Input:** A list of {N} raw text objects, each containing the title, abstract, and a unique ID for a publication.

**Output:** A single JSON object containing a list of {K} thematic clusters, where each cluster has a title and a list of associated publication IDs.

This structured output is critical for the automation of the subsequent stages, serving as an intermediate data structure that is passed to the next stage of the pipeline without being exposed to the end-user.

### B. Stage 2: Intra-Cluster Trend Synthesis

*Objective:* With publications now grouped thematically, the next stage is to synthesize the key findings and narrative within each theme. The goal is to transform a simple list of papers into a coherent, summary paragraph that explains what is currently happening in that specific sub-field.

*Methodology:* For each thematic cluster generated in Stage 1, a second, more focused prompt is executed. This prompt instructs the LLM to act as a scientific writer. Crucially, it is constrained to synthesize a narrative rather than simply listing the findings of each paper. This methodological choice is designed to leverage the inherent few-shot learning capabilities of large-scale models, which enable them to understand and generate nuanced human language far beyond the scope of simple keyword matching or extractive summarization [11].

**Prompt 2 (Pseudo-code):**

ROLE: You are a skilled scientific writer for a prestigious journal.

CONTEXT: You are provided with the titles and abstracts for a cluster of papers all focused on the theme of: "{Theme Title}".

TASK:

1. Read and analyze all provided texts within this theme.

2. Write a single, concise, and insightful paragraph (approx. 150 words) that summarizes the key trends, findings, and debates within this topic.

CONSTRAINT: Do not simply list what each paper did. Instead, synthesize a coherent narrative. For example, start with a topic sentence and then elaborate on the common direction or conflicting results presented in the papers.

**Input:** The JSON object from Stage 1, processed one theme at a time.

**Output:** A set of {K} text paragraphs, each representing a synthesized summary of a single research theme.

### C. Stage 3: Section Generation

*Objective:* The final stage is to assemble the individual thematic summaries from Stage 2 into a single, cohesive, and professionally structured section of the digest. The goal is to move beyond a simple concatenation of paragraphs and generate a human-readable section that includes a high-level description of the detailed trends and a forward-looking conclusion.

*Methodology:* This is accomplished with a single, complex "master prompt" that guides the LLM to perform multiple cognitive tasks within one generative cycle. The prompt instructs the model to adopt the persona of an expert editor and to structure its output into distinct sub-components. This structured prompting, which provides the model with pre-synthesized and highly relevant context, is a deliberate methodological choice designed to mitigate the risk of factual "hallucinations" - a widely acknowledged challenge in generative models [3].

**Prompt 3 (Pseudo-code):**

ROLE: You are an expert research editor compiling a "Key Trends" section for a prestigious scientific digest.

CONTEXT: You are provided with {K} distinct summary paragraphs, each detailing a key research trend in [Domain] from the last month.

TASK: Your task is to write a complete, self-contained "Key Trends" section based on the provided information. The section must have the following two parts:

1. Main Body: Present each of the {K} summary paragraphs. You may add short transitional phrases between them to ensure the text flows logically, but preserve the core information of each summary.

2. Concluding Analysis: After presenting the trends, write a final, short paragraph (2-3 sentences) titled "Concluding Analysis". In this part, do not repeat the summaries. Instead, provide a brief, forward-looking synthesis. Answer the question: "Based on these trends, what is the most likely future direction of research in this field?"

OUTPUT FORMAT: Return the final text as a single, ready-to-publish section with the two distinct parts clearly separated.

**Input:** The set of {K} text paragraphs from Stage 2.

**Output:** A single, structured block of text representing a complete analytical section (e.g., the "Key Trends" section), internally divided into a main body and a concluding analysis. This process is repeated with different master prompts to generate all necessary sections for the report. Finally, these generated text blocks are programmatically assembled to form the single, complete monthly digest for the specified domain, such as Education, Finance and Medicine.

In summary, the three-stage prompt chaining pipeline transforms the complex task of trend analysis into a deterministic, step-by-step workflow. By systematically converting a large volume of unstructured texts into thematic clusters, then into synthesized narratives, and finally into structured report sections, this methodology provides a high degree of control over the LLM's output. This structured approach is designed not only to enhance the quality of the final digest but also to ensure the process is repeatable and scalable. Having detailed the architecture of this pipeline, we now turn to its quantitative evaluation to validate its effectiveness in practice. The following section will present the experimental setup and performance results.

## IV. Experimental Setup

To validate the effectiveness and reliability of our proposed prompt chaining pipeline, we designed a comparative experiment. The objective was to quantitatively measure the performance of our multi-stage approach against the strongest possible version of a single-shot prompting baseline, using a human-authored "gold standard" report as the ground truth. The methodology of our comparative experiment is depicted in **Figure 2**.

### A. Research Questions

Our experiment was designed to answer the following key questions:

*1) Quality:* Does the prompt chaining approach produce a final digest that is qualitatively closer to a human-authored report compared to an optimized single-shot prompt?

*2) Reliability:* Even when highly optimized, is the single-shot prompt less consistent and more prone to catastrophic failures than the prompt chaining approach?

### B. Systems Under Comparison

Three distinct versions of the digest were created for the experiment:.

*1) Ground Truth (Gold Standard):* A human expert with a Doctorate in Pedagogy was engaged to manually review the dataset and write a "gold standard" digest. This report served as the ideal target output.

*2) Baseline (Optimized Single-Shot Prompting):* To ensure a fair and challenging comparison, we deliberately avoided a simplistic, naive baseline. Instead, we developed a "strong baseline" through an iterative process of prompt engineering. We constructed a single, monolithic prompt that logically embeds all the tasks of our three-stage chain. This prompt was refined over several iterations to be as clear and unambiguous as possible, representing the strongest possible version of a single-shot approach we could engineer. The pseudo-code for this prompt is shown below.

**Baseline Prompt (Pseudo-code):**

ROLE: You are an expert research editor and analyst compiling a "Key findings and research outcomes" section for a prestigious scientific digest.

CONTEXT: You are provided with a list of {N} academic paper titles and abstracts related to the field of Finance. Your final output must be a single, cohesive, and ready-to-publish analytical section.

TASK: Execute the following sequence of cognitive tasks IN A SINGLE PASS to generate the final output:

1. Thematic Analysis: Read and analyze all provided texts. Internally identify the {K=5} most dominant and distinct research themes.

2. Intra-Theme Synthesis: For each of the {K} themes you identified, internally formulate a concise summary paragraph (approx. 200 words) that captures its significant results and methodologies.

3. Final Section Generation: Based on your internal analysis from the previous steps, write the complete, self-contained "Key findings and research outcomes" section. This section must synthesize all themes together as a whole.

CONSTRAINTS & OUTPUT FORMAT:

1. DO NOT provide a breakdown for each individual topic in the final output. The final text must be a single, integrated analysis.

2. CITE specific findings, trends, or comparisons by referencing the original article's Title and DOI.

3. The final output must be ONLY the fully generated section, ready for publication. Do not include any preliminary notes, lists of themes, or other intermediate artifacts.

*3) Proposed Method (Prompt Chaining):* This is the AI SciBrief system as described in Section III, which utilizes the three-stage prompt chaining pipeline.

### C. Dataset and Procedure

The AI SciBrief platform is designed to be domain-agnostic. For this experiment, we selected the "Education" domain to align with the expertise of our evaluating expert. The experiment was conducted on the set of all English-language publications from this domain indexed by OpenAlex for the month of July 2025.

The experimental procedure was as follows:

*1)* The human expert authored the **Gold Standard** digest.

*2)* **The Proposed Method (Prompt Chaining)** was executed once to generate its version of the digest.

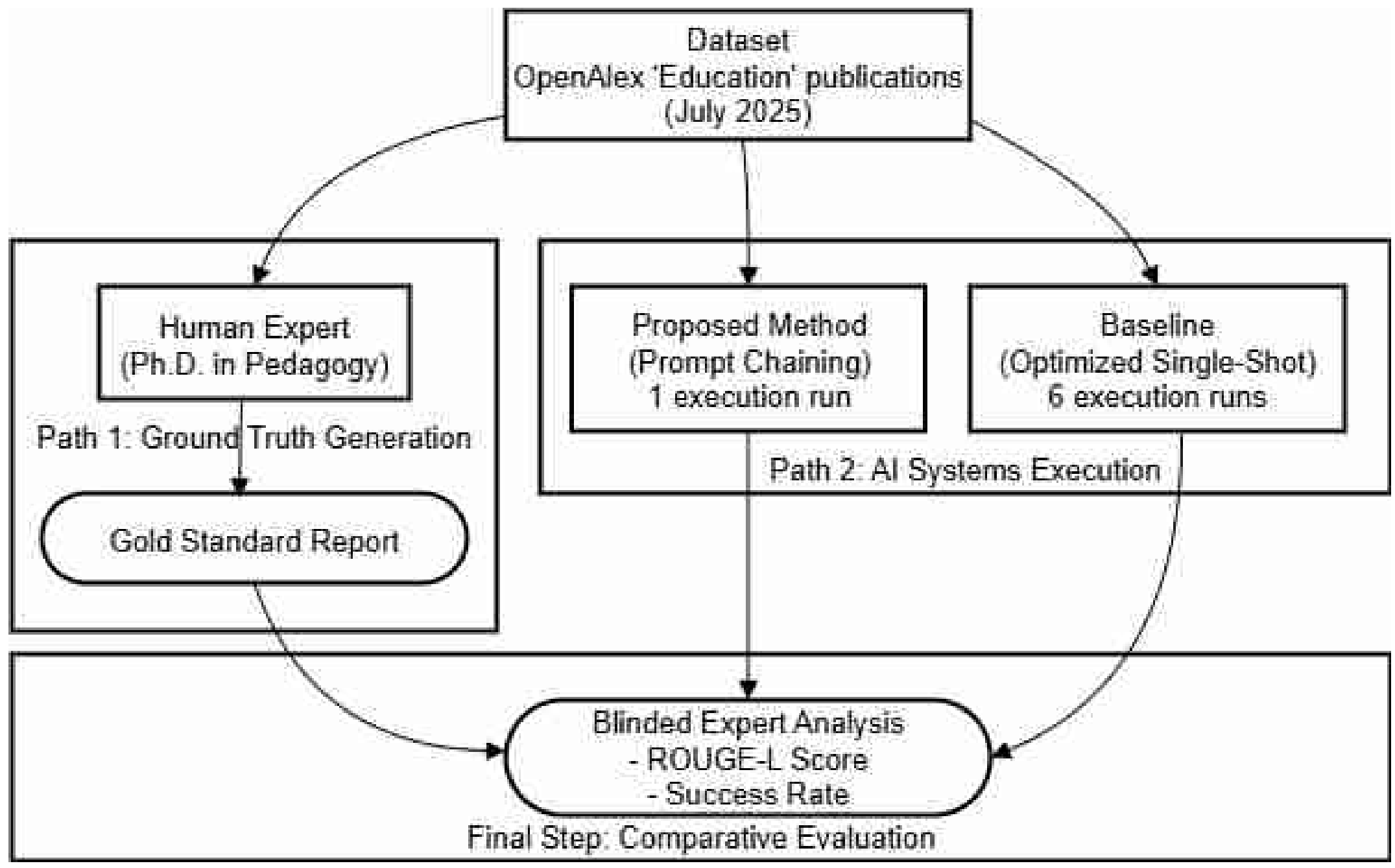

Fig. 2. The methodology of the comparative experiment

*3)* To test for reliability and consistency, the **Baseline (Optimized Single-Shot Prompting)** was executed **six times** on the identical dataset. This was done to assess the variability and failure rate of even a highly refined, complex single-shot instruction.

### *D. Evaluation Metrics*

We employed both quantitative and qualitative metrics for evaluation.

*1) Quantitative (Quality):* To provide a nuanced evaluation of text quality, we used the **ROUGE-L** metric, which measures the longest common subsequence between the generated text and a reference, thereby rewarding structural similarity. We report three scores from the ROUGE-L calculation:

*a) Precision:* This metric indicates the proportion of the generated text that is also found in the gold standard. A high precision score suggests that the system's output is focused and contains minimal irrelevant or "hallucinated" content.

*b) Recall:* This metric measures the proportion of the gold standard text that is captured by the generated output. A high recall score indicates that the system successfully covered most of the essential information.

*c) F1-score:* The harmonic mean of Precision and Recall, providing a single, balanced measure of overall quality.

The human expert, who authored the Gold Standard, scored the outputs of both the Proposed Method and the successful runs of the Baseline against his own report. To prevent bias, the expert was blinded as to which system produced which output.

*2) Qualitative (Reliability):* For the six runs of the Baseline system, we measured the Success Rate. An output was deemed a "Failure" if it met one of two conditions:

*a)* It failed to produce a structured, coherent final section, instead returning only the intermediate thematic groupings, or

*b)* The generated content was critically off-topic.

The results of this experiment are presented and discussed in the following section.

## V. Results and Discussion

This section presents the results of our comparative experiment, designed to evaluate the quality and reliability of the prompt chaining approach against an optimized single-shot baseline. We first present the quantitative results from our evaluation metrics, followed by a qualitative analysis of the baseline system's failure modes. Finally, we discuss the implications of these findings in the context of our research questions.

### *A. Quantitative Results*

The performance of the two systems was measured against the human-authored Gold Standard report. The ROUGE-L metric was used to evaluate quality, with F1-score providing a balanced view, while Precision and Recall offer deeper insights into the nature of the generated text. The key results are summarized in **Table 1**.

To better visualize the quality difference, the ROUGE-L scores are presented as a bar chart in **Figure 3**.

The quantitative results reveal three critical findings:

*1) Reliability:* There is a significant difference in reliability. The Proposed Method succeeded on its first and only attempt. The optimized Single-Shot Baseline, however, failed in 50% of its execution runs, demonstrating significant inconsistency.

*2)* Overall Quality (F1-score): In terms of balanced quality, the Prompt Chaining method (F1=0.507) clearly outperformed the average successful run of the Baseline (F1=0.486), indicating its output was structurally and semantically closer to the human-authored Gold Standard.

TABLE I. COMPARATIVE PERFORMANCE OF PROMPT CHAINING VS. SINGLE-SHOT BASELINE

| System Under Test | Execution Runs | Success Rate | ROUGE-L Precision | ROUGE-L Recall | ROUGE-L F1-score |
|---|---|---|---|---|---|
| Proposed Method (Prompt Chaining) | 1 | 100% (1/1) | 0.535 | 0.482 | 0.507 |
| Baseline (Optimized Single-Shot) | 6 | 50% (3/6) | 0.510[a] | 0.464[a] | 0.486 |

[a.] Note: ROUGE scores for the Baseline are averaged across the 3 successful runs.

*3) Nature of the Output (Precision vs. Recall):* The breakdown into Precision and Recall is particularly revealing. **The Proposed Method** achieved a high Precision score (0.535), indicating that the majority of the information it included was relevant and aligned with the Gold Standard. Its Recall (0.482) was also strong. In contrast, **the Baseline** showed significantly lower Precision (0.510), suggesting it was more prone to including extraneous, less relevant, or slightly off-topic information ("hallucinations"). Its Recall (0.464) was comparable to our method, indicating it was able to find most of the important information, but struggled to present it concisely and accurately.

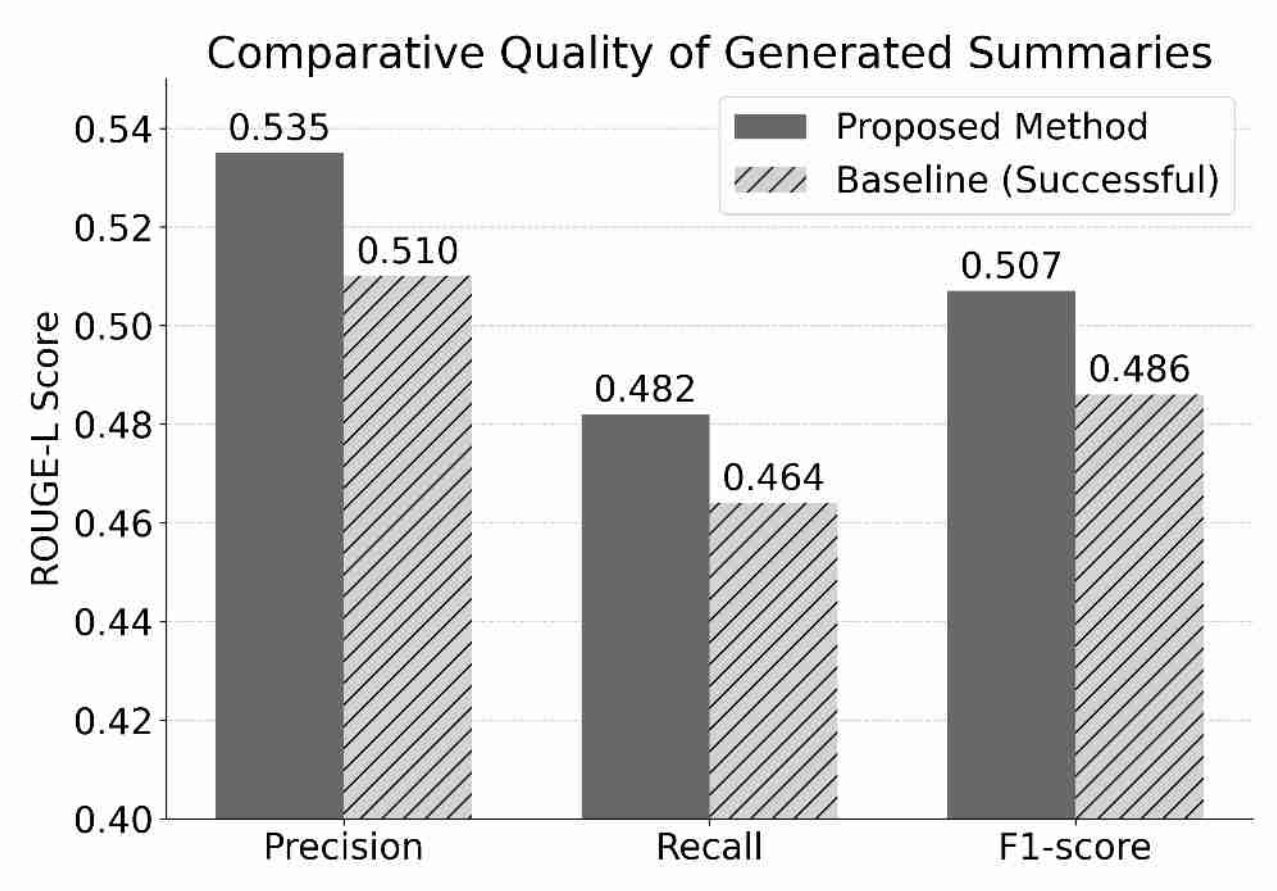

Fig. 3. ROUGE-L scores of the Proposed Method versus the successful runs of the Baseline.

## B. Qualitative Analysis of Baseline Failures

A qualitative review of the three failed runs of the Baseline system revealed two distinct failure modes:

*1) Structural Failure (2 out of 3 failures):* In two cases, the LLM successfully executed the initial parts of the monolithic prompt (identifying and grouping articles into five themes) but failed to complete the final synthesis step. The output was an unstructured collection of the five thematic summaries without the required narrative assembly or concluding analysis. This suggests the model failed to maintain the instructional context of the complex sequence of instructions towards the end of the generative process.

*2) Topical Deviation (1 out of 3 failures):* In one case, the system produced a well-structured but critically off-topic report. While the text was related to "education" in a very broad sense, it completely missed the specific nuances and trends present in the source articles for July 2025. This represents a catastrophic failure, as the output is confidently incorrect and misleading, which is more dangerous for an automated report than no output at all.

## C. Discussion

The experimental results provide clear answers to our initial research questions.

In response to **RQ1 (Quality)**, our findings indicate that the prompt chaining approach yields a qualitatively superior output. The higher F1-score is primarily driven by a significant improvement in **precision**. This suggests that by deconstructing the task, each stage of our pipeline operates with a more focused context, which helps the LLM to avoid generating irrelevant or redundant content and to adhere more closely to the source material. The Baseline, tasked with a more complex, monolithic instruction, struggled to maintain this focus, resulting in a less precise summary.

The answer to **RQ2 (Reliability)** is even more definitive. The 50% failure rate of the strong baseline, despite its optimization, strongly supports our central hypothesis. We theorize that this is due to a form of "cognitive overload" or "instruction interference" within the LLM. While modern LLMs are powerful, providing them with a long and complex sequence of instructions in a single prompt increases the probability of error, as the model may fail to correctly weight and execute all constraints.

Our multi-stage approach directly mitigates this risk. It operationalizes the core principle behind foundational techniques like Chain-of-Thought (CoT) prompting, that breaking down a problem enhances reasoning, and applies it at a workflow level. By creating a pipeline where each prompt has a single, well-defined task, we reduce the cognitive load on the model at each step, dramatically increasing the reliability of the end-to-end process.

The broader implication of this finding is significant for the field of applied AI. For complex, multi-step generative tasks, prompt chaining should be considered not merely an alternative, but a more robust and reliable architectural pattern than monolithic prompting. This case study demonstrates that engineering the **process** of interaction with an LLM is as important as engineering the **prompts** themselves.

# VI. CONCLUSION AND FUTURE WORK

## A. Conclusion

This paper presented a practical implementation and rigorous empirical evaluation of a prompt chaining methodology for the automated generation of scholarly digests. Our central contribution is the demonstration that a multi-stage architectural pattern is not merely an alternative to a monolithic, single-shot prompt, but a fundamentally more robust and reliable approach for complex generative tasks.

Our experiments provided clear, quantitative evidence to support this conclusion. The proposed three-stage pipeline, implemented in the AI SciBrief system, achieved a 100% success rate in our tests. In contrast, a carefully optimized single-shot baseline proved to be highly unreliable, failing in 50% of its execution runs. Furthermore, even in its successful attempts, the baseline's output was qualitatively inferior, exhibiting lower precision as measured by the ROUGE-L metric. These findings suggest that by deconstructing a complex task into a sequence of focused, manageable steps, prompt chaining significantly reduces the risk of cognitive overload and instruction interference in the LLM, leading to more consistent and higher-quality results.

Ultimately, this work provides a validated methodology for engineers and researchers seeking to build dependable, automated systems for knowledge synthesis. We have shown that the path to reliable AI-driven analysis lies not just in the sophistication of the models themselves, but in the deliberate, structured engineering of the human-to-AI interaction process.

### B. Future Work

The results presented in this study clearly demonstrate the primary advantage of our proposed Prompt Chaining method: superior reliability and robustness. With a 100% success rate, our method effectively overcomes the brittleness of the Optimized Single-Shot baseline, which fails in 50% of cases due to an inability to adhere to formal output constraints. This is a significant step towards creating more dependable and predictable generative systems.

However, a closer analysis of the ROUGE-L scores reveals a more nuanced picture. While our method outperforms the baseline's average, the margin is modest when compared specifically to the baseline's successful runs (an F1-score of 0.507 vs. 0.486). This indicates that our current advantage stems largely from consistency and eliminating failures, rather than generating a qualitatively superior output in every instance.

Therefore, the primary direction for our future work will be to enhance the intrinsic quality of the output generated by our chain pipeline. The goal is to evolve the method so that it not only maintains its perfect success rate but also significantly surpasses the quality of the baseline's successful outputs.

To achieve this, we plan to focus on the following key areas:

*1) Optimization of Individual Chain Links:* We will conduct a granular analysis of each prompt within the chain. By refining the instructions, examples, and constraints for each sub-task, we aim to improve the quality of intermediate outputs, which should cumulatively lead to a better final result.

*2) Enhanced Context Propagation:* We will explore more sophisticated techniques for passing context between the links in the chain. This includes experimenting with dynamic context summarization and explicitly instructing later steps to review, verify, or build upon the information generated in earlier ones, thereby improving overall coherence and accuracy.

## ACKNOWLEDGMENT

The author wishes to express his sincere gratitude to Dr. Aleksandr A. Rusakov, President of the Academy of Informatization of Education, for his invaluable contribution in developing the expert-authored "gold standard" report that served as the benchmark for this study. His academic rigor and expertise were instrumental in ensuring the objectivity and validity of our experimental results.